\documentclass[twoside,11pt]{article}

\usepackage{amsmath,amsfonts,bm}

\def\eqref#1{equation~\ref{#1}}

\def\floor#1{\lfloor #1 \rfloor}
\def\1{\bm{1}}

\DeclareMathAlphabet{\mathsfit}{\encodingdefault}{\sfdefault}{m}{sl}
\SetMathAlphabet{\mathsfit}{bold}{\encodingdefault}{\sfdefault}{bx}{n}

\usepackage{jmlr2e}
\usepackage{url}
\usepackage{xcolor}
\usepackage{amsmath}
\usepackage[format=plain]{caption}
\usepackage{float}
\usepackage{bbold}
\usepackage{mathtools}
\usepackage[ruled]{algorithm2e}
\usepackage{multirow}
\usepackage{wrapfig}
\usepackage{booktabs}

\ShortHeadings{Effective Sparsity in Neural Networks}{Vysogorets and Kempe}
\firstpageno{1}

\usepackage{lastpage}
\jmlrheading{24}{2023}{1-\pageref{LastPage}}{4/22; Revised
2/23}{3/23}{22-0415}{Artem Vysogorets, Julia Kempe}
\ShortHeadings{Pruning Through the Lens of Effective Sparsity}{Vysogorets \& Kempe}

\begin{document}
\title{Connectivity Matters: Neural Network Pruning Through the Lens of Effective Sparsity}
\author{\name Artem Vysogorets \email amv458@nyu.edu\\
       \addr Center for Data Science \\
       New York University\\
       New York, NY 10011, USA
       \AND
       \name Julia Kempe \email jk185@nyu.edu \\
       \addr Center for Data Science\\
       Courant Institute for Mathematical Sciences\\
       New York University\\
       New York, NY 10011, USA}
\editor{Pradeep Ravikumar}

\maketitle 

\begin{abstract}
Neural network pruning is a fruitful area of research with surging interest in high sparsity regimes. Benchmarking in this domain heavily relies on faithful representation of the sparsity of subnetworks, which has been traditionally computed as the fraction of removed connections (direct sparsity). This definition, however, fails to recognize unpruned parameters that detached from input or output layers of the underlying subnetworks, potentially underestimating actual effective sparsity: the fraction of inactivated connections. While this effect might be negligible for moderately pruned networks (up to $10\times$--$100\times$ compression rates), we find that it plays an increasing role for sparser subnetworks, greatly distorting comparison between different pruning algorithms. For example, we show that effective compression of a randomly pruned LeNet-300-100 can be orders of magnitude larger than its direct counterpart, while no discrepancy is ever observed when using SynFlow for pruning \citep{synflow}. In this work, we adopt the lens of effective sparsity to reevaluate several recent pruning algorithms on common benchmark architectures (e.g., LeNet-300-100, VGG-19, ResNet-18) and discover that their absolute and relative performance changes dramatically in this new, and as we argue, more appropriate framework. To aim for effective, rather than direct, sparsity, we develop a low-cost extension to most pruning algorithms. Further, equipped with effective sparsity as a reference frame, we partially reconfirm that random pruning with appropriate sparsity allocation across layers performs as well or better than more sophisticated algorithms for pruning at initialization \citep{sanitychecks}. In response to this observation, using an analogy of pressure distribution in coupled cylinders from thermodynamics, we design novel layerwise sparsity quotas that outperform all existing baselines in the context of random pruning.
\end{abstract}


\section{Introduction}
Recent successful advances of Deep Neural Networks are commonly attributed to their high architectural complexity and excessive size (\emph{over-parameterization}) \citep{denton,neyshabur,arora}. Modern state-of-the-art architectures exhibit enormous parameter overhead, requiring prohibitive amounts of resources during both training and inference and leaving a significant environmental footprint \citep{megatron}. In response to these challenges, much attention has turned to compression of neural networks and, in particular, parameter pruning. While initial approaches mostly focused on pruning models after training \citep{obd,obs}, contemporary algorithms optimize the sparsity structure of a network while training its parameters \citep{set,rigl} or even remove connections before any training whatsoever \citep{snip,grasp}.

Compression rates considered in the pruning literature usually fall between $10\times$ and $100\times$ of the size of the original model. However, as contemporary model sizes grow into the billions of parameters, studying higher compression regimes becomes increasingly important.  Recently, a new bold sparsity benchmark was set by \citet{synflow} with Iterative Synaptic Flow (SynFlow), a data-agnostic algorithm for pruning at initialization. Reportedly, it is capable of removing all but only a few hundreds of parameters (a $100,000\times$ compression for VGG-16) and still produce trainable subnetworks, while other pruning methods disconnect networks at much lower sparsity levels \citep{synflow}. Related work by \citet{force} proposes an iterative version of one-shot pruning algorithm, Single-shot Network Pruning (SNIP) \citep{snip}, and evaluates it in a similar high sparsity regime, reaching more than $10,000\times$ compression ratio.

\begin{figure}[H]
\begin{center}
\includegraphics[width=0.3\linewidth]{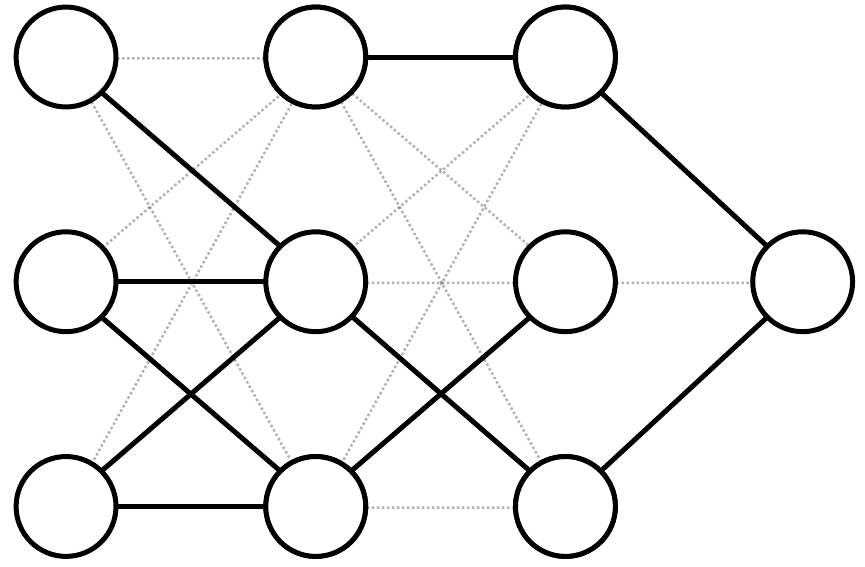}\qquad\qquad
\includegraphics[width=0.3\linewidth]{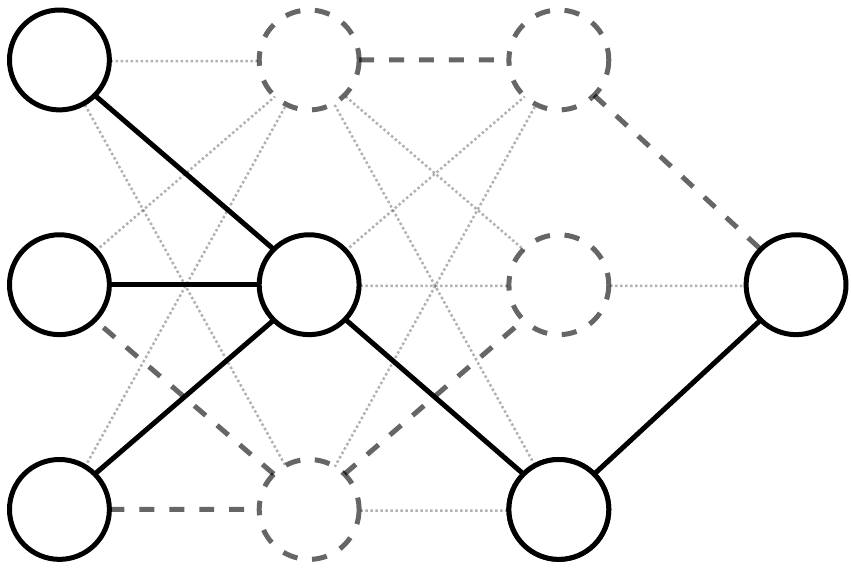}
\end{center}
\caption{Pruning $11$ edges from a fully-connected $21$-edge network. Left: direct sparsity ($11/21$) does not account for disconnected edges (compression $21/10=2.1$). Right: effective sparsity ($16/21$) accounts for the $5$ dashed connections incident to inactivated neurons (yielding twice as large effective compression $21/5=4.2$).}
\label{Fig:1}
\end{figure}

\paragraph{Effective sparsity.} This increased focus on extreme sparsity leads us to consider \emph{what sparsity is meant to represent} in neural networks and computational graphs at large. In the context of neural network pruning, sparsity to date is computed straightforwardly as the \emph{fraction of removed connections} (\emph{direct sparsity})---and compression as the inverse fraction of unpruned connections (\emph{direct compression}). We observe that this definition does not distinguish between connections that have actually been pruned, and those that have become \emph{effectively} pruned because they have disconnected from the computational flow. Formally, an edge $\theta_i$ is considered inactive if, for any input $x$, the output of the neural network $f(\theta,x)$ does not depend on the value of $\theta_i$. In this work, we propose to abandon direct sparsity in favor of \emph{effective sparsity}---the \emph{fraction of inactivated connections}, be it through direct pruning or through otherwise disconnecting from either input or output of a network (see Figure \ref{Fig:1} for an illustration).

\begin{figure}[h]
\centering
\includegraphics[width=0.49\linewidth]{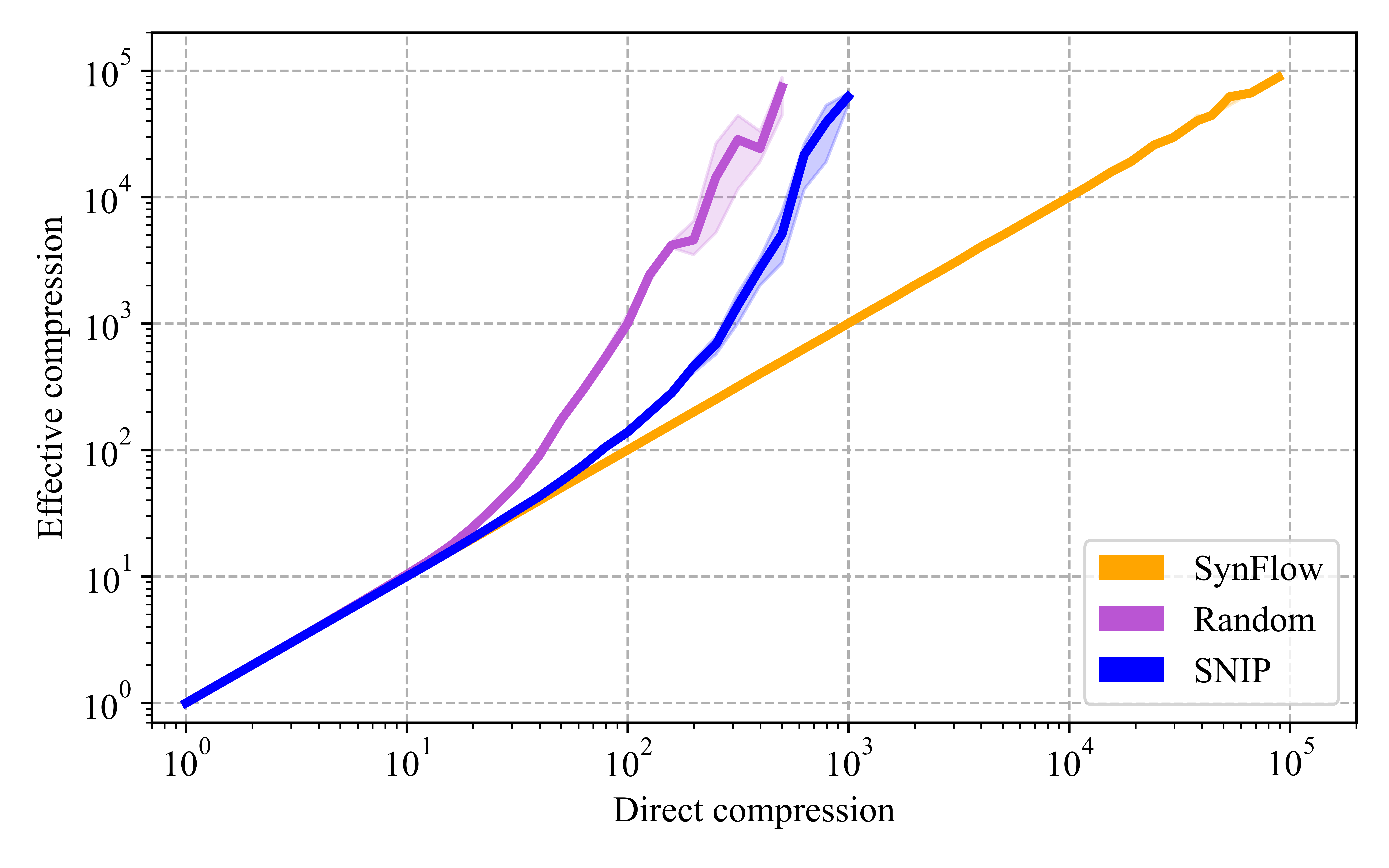}
\includegraphics[width=0.49\linewidth]{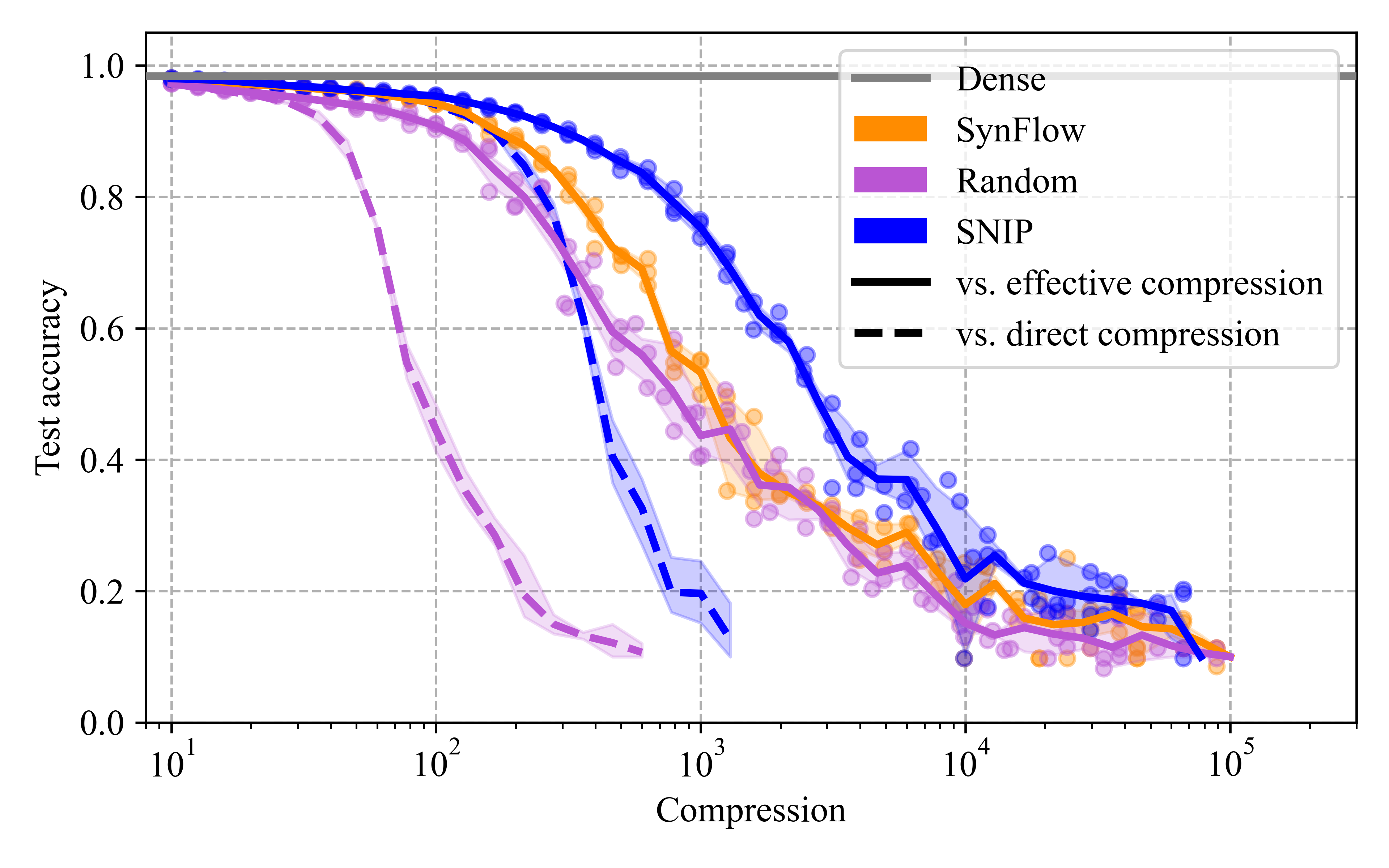}
\caption{LeNet-300-100 trained on MNIST after pruning. Left: gaps between direct and effective compression. Right: SynFlow has a better sparsity-accuracy trade-off than SNIP when plotted against direct (dashed), but not against effective compression (solid curves). Dots represent individual experiments. Dashed and solid curves coincide for SynFlow.}
\label{Fig:2}
\end{figure}

We advocate that effective sparsity (effective compression) be used universally in place of its direct counterpart since it more accurately depicts what one would reasonably consider the network's sparsity state. Using the lens of effective compression for benchmarking  allows for a fairer comparison between different unstructured pruning algorithms. Note that effective compression is lower bounded by direct compression, which means that some pruning algorithms will give improved sparsity-accuracy trade-offs in this new framework. In Section \ref{Sec:EffectiveSparsity}, we critically reexamine a plethora of recent pruning algorithms for a variety of architectures to find that, in this refined framework, conclusions drawn in previous works appear overstated or incorrect. Figure \ref{Fig:2} gives a sneak-preview of this effect for three ab-initio pruning algorithms: SynFlow \citep{synflow}, SNIP \citep{snip} and plain random pruning for LeNet-300-100 on MNIST. While SynFlow appears superior to other methods when evaluated against direct compression, it loses its advantage in the effective framework. 
Such radical performance changes are partly explained by differing gaps between effective and direct compression inherent to different pruning algorithms (Figure \ref{Fig:2}). We can see that significant departure of direct from effective compression kicks in at relatively low rates below $100\times$, making our work relevant even in these moderate regimes. For example, using random pruning to compress LeNet-300-100 by $100\times$ (sparsity $99\%$) results in $\sim 1,000\times$ effective compression; yet, removing the same number of parameters with SynFlow yields an unchanged $100\times$ effective compression.  What makes certain iterative algorithms like SynFlow less likely to amass disconnected edges? In Section \ref{Sec:EffectiveSparsity}, we show that they are fortuitously designed to achieve a close convergence of direct and effective sparsity, hinting
  that preserving connectivity is an important aspect in the strong performance of high-compression pruning algorithms \citep{synflow,force}. Moreover, the lens of effective compression gives access to more extreme compression regimes for some pruning algorithms, which appear to disconnect much earlier when not accounting for inactive connections. For these high effective compression ratios all three pruning methods from Figure \ref{Fig:2} perform surprisingly similar, even though they use varying degrees of information on data and parameter values.

\paragraph{Layerwise Sparsity Quotas (LSQ) and Ideal Gas Quotas (IGQ).} A recent thread of research by \citet{missingthemark} and \citet{sanitychecks} shows that performance of trained subnetworks produced by algorithms for pruning at initialization is robust to randomly reshuffling unpruned edges within layers before training. This observation led to the conjecture that these algorithms essentially generate successful distributions of sparsity across layers, while the exact connectivity patterns are unimportant. In Section \ref{Sec:Pistons}, we reexamine this conjecture through the lens of effective sparsity, confirm it for moderate compression regimes ($10\times$--$100\times$) studied by \citet{missingthemark} and \citet{sanitychecks}, but find the truth to be more nuanced at higher compression rates. Nonetheless, this result highlights the importance of algorithms that carefully engineer \emph{layerwise sparsity quotas (LSQ)} to obtain very simple and adequately performing random pruning algorithms. Furthermore, concurrently with our work, \citet{unreasonable}, find that random pruning with carefully allocated sparsity among layers can match the performance of their dense counterparts. Another important motivation to search for good LSQ is that global pruning algorithms frequently remove entire layers prematurely \citep{snip2} (cf. layer-collapse in  \citet{synflow}), even before any significant differences between direct and effective sparsity emerge. Well-engineered LSQ could avoid this and enforce proper redistribution of compression across layers (see \citet{gale,set,rigl} for existing baselines). In Section \ref{Sec:Pistons}, we propose a novel LSQ coined \emph{Ideal Gas Quotas (IGQ)} by drawing intuitive analogies from physics. Effortlessly computable for any network-sparsity combination, IGQ performs similarly or better than any other baseline in the context of random pruning at initialization and of magnitude pruning after training.

\paragraph{Effective pruning.} Pruning to any desired direct sparsity is straightforward: one simply needs to mask out the corresponding number of parameters from a network. Effective sparsity, unfortunately, is more unpredictable and difficult to control. In particular, several known pruning algorithms suffer from layer-collapse once reaching a certain sparsity level, leading to unstable effective sparsity just before the disconnection. As a result, most pruning methods are unable to deliver certain values of effective sparsity regardless of how many connections are pruned. When possible, however, one needs to carefully tune the number of pruned parameters so that effective sparsity lands near a desired value. In Section \ref{Sec:Pruner}, we suggest a simple extension to algorithms for pruning at initialization or after training that helps bring effective sparsity close to any predefined achievable value while incurring costs that are at most logarithmic in model size.

\paragraph{Our contributions: summary.} In this study, we $(i)$ formulate and illustrate the importance of effective sparsity by reevaluating several recent pruning strategies; $(ii)$ provide algorithms to prune according to and compute effective sparsity; $(iii)$ reconfirm that networks pruned at initialization are robust to layerwise reshuffling of survived edges  \citep{missingthemark} in the new sparsity framework, and (iv) design efficient layerwise sparsity quotas IGQ for random pruning that perform consistently well across all sparsity regimes.

\section{Related work}
\label{Sec:RelatedWork}
Neural network compression encompasses a number of orthogonal approaches such as parameter regularization \citep{reg2,reg}, variational dropout \citep{dropout}, vector quantization and parameter sharing \citep{quant,hash,quant2}, low-rank matrix decomposition \citep{denton,lowrank}, and knowledge distillation \citep{caruana,distillation}. Network pruning, however, is by far the most common technique for model compression, and can be partitioned into structured (at the level of entire neurons/units) and unstructured (at the level of individual connections). While the former offers resource efficiency unconditioned on use of specialized hardware \citep{hardware} and constitutes a fruitful research area \citep{structured1,structured2}, we focus on the more actively studied unstructured pruning, which is where differences between effective and direct sparsity emerge. In what follows we give a quick overview, naturally grouping pruning methods by the time they are applied relative to training (see \citet{lth} and \citet{grasp} for a similar taxonomy).

\paragraph{Pruning after training.} These earliest pruning techniques were designed to remove the least ``salient'' learned connections without sacrificing predictive performance. Optimal Brain Damage \citep{obd} and its sequel Optimal Brain Surgeon \citep{obs} use the Hessian of the loss to estimate sensitivity to removal of individual parameters. \citet{han} popularized magnitude as a simple and effective pruning criterion. It proved to be especially successful when applied alternately with several finetuning cycles, which is commonly referred to as Iterative Magnitude Pruning (IMP), a modification of which was used by \citet{lth} to discover lottery tickets---sparse subnetworks that achieve the performance of their dense counterparts within a commensurate number of iterations. Later, \citet{distortion1} showed that magnitude-based pruning minimizes $\ell_2$ distortion of each layer's output incurred by parameter removal. Recently, \citet{lamp} extend this idea and propose Layer-Adaptive Magnitude-Based Pruning (LAMP), which approximately minimizes the upper bound of the $\ell_2$ distortion of the entire network. While equivalent to magnitude pruning within individual layers, LAMP automatically discovers excellent layerwise sparsity quotas (see Section \ref{Sec:Pistons}) that yield better performance (as a function of \emph{direct} compression) than existing alternatives in the context of IMP. 

\paragraph{Pruning during training.} Algorithms in this category learn sparsity structures together with parameter values, hoping  that continued training will correct for damage incurred by pruning. To avoid inefficient prune-retrain cycles inherent to IMP, \citet{narang} introduce gradual magnitude pruning over a single training round. Subsequently, \citet{zhugupta} modify this algorithm by introducing a simpler pruning schedule and keeping layerwise sparsities uniform throughout training. Sparse Evolutionary Training (SET) \citep{set} starts with an already sparse subnetwork and restructures it during training by pruning and randomly reviving connections. Unlike SET, \citet{dsr} allow redistribution of sparsity across layers, while \citet{snsf} use gradient momentum as the criterion for parameter regrowth. \citet{rigl} rely on the instantaneous gradient to revive weights but follow SET to maintain the initial layerwise sparsity distribution during training. The In-time Over-parameterization (ITOP) framework provides insights into the underlying mechanisms of the above methods and leads to improved training protocols that boost their performance \citep{itop}. A different body of works tackle the general optimization problem with an intractable $\ell_0$ parameter sparsity constraint by designing and solving related continuous problems \citep{probmask,savarese,str}. For example, Continuous Sparsification (CS) by \citet{savarese} uses a sigmoid of learnable continuous variables as mask values and applies $\ell_1$ regularization, effectively forcing them to either $0$ or $1$ during training.

\paragraph{Pruning before training.} Pruning at initialization is especially alluring to deep learning practitioners as it promises lower costs of both optimization and inference. While this may seem too ambitious, the Lottery Ticket Hypothesis (LTH) postulates that randomly initialized dense networks do indeed contain highly trainable and equally well-performing sparse subnetworks \citep{lth}. Inspired by the LTH, \citet{snip} design SNIP, which uses connection sensitivity as a parameter saliency score. \citet{grasp} notice that SNIP creates bottlenecks or even removes entire layers and propose Gradient Signal Preservation (GraSP) as an alternative that aims to maximize gradient flow in a pruned network. \citet{force} improve SNIP by applying it iteratively, allowing for reassessment of saliency scores during pruning and helping networks stay connected at higher compression rates. A truly new compression benchmark was set by \citet{synflow}; their algorithm, SynFlow, iteratively prunes subsets of parameters according to their $\ell_1$ path norm and helps networks reach maximum compression without disconnecting. For example, SynFlow achieves non-random test accuracy on CIFAR-10 with a $100,000\times$ compressed VGG-16, while SNIP and GraSP fail already at $100\times$ and $1,000\times$, respectively. An extensive ablation study by \citet{missingthemark} examines SNIP, GraSP and SynFlow within moderate compression rates (up to $100\times$) and reveals that performance of subnetworks produced by these methods is stable under random layerwise rearrangement of edges prior to training. Later, this result was independently confirmed by \citet{sanitychecks} for SNIP and GraSP only. This observation suggests that these algorithms perform as well as random pruning with corresponding layerwise quotas, putting the spotlight on designing competitive LSQ \citep{set,gale,lamp}. \citet{ilan} augment the functionality of sparse layers by precomputing a deterministic transformation of the input, thus maintaining information propagation and avoiding layer-collapse.

\section{Effective sparsity}
\label{Sec:EffectiveSparsity}
In this section, we present our comparisons of a variety of pruning algorithms under the lens of effective compression. To illustrate the striking difference between direct and effective sparsity and expose the often radical change in relative performance of pruning algorithms when switching from the former to the latter, we evaluate several recent methods (SNIP, GraSP, SynFlow, Magnitude \& LAMP\footnote{as two versions of magnitude pruning after training and close siblings of lottery tickets \citep{lth}.}, CS\footnote{as a representative of methods that use learnable sparsity}, SNIP-iterative) and random pruning with uniform sparsity distribution across layers in both frameworks. Our experiments encompass modern architectures on commonly used computer vision benchmark datasets: LeNet-300-100 \citep{lenets} on MNIST, LeNet-5 \citep{lenets} on CIFAR-10, VGG-19 \citep{vggnets} on CIFAR-100, ResNet-18 \citep{resnet} on TinyImageNet, and ResNet-50 and MobileNetV2 \citep{mobilenets} on ImageNet. We place results of VGG-16 \citep{vggnets} on CIFAR-10 in Appendix \ref{Sec:AppendixVGG16}, as they closely resemble those of VGG-19. Further experimental details are listed in Appendix \ref{Sec:AppendixHyperparameters}.

\begin{figure}[H]
\centering
\includegraphics[width=0.8\linewidth]{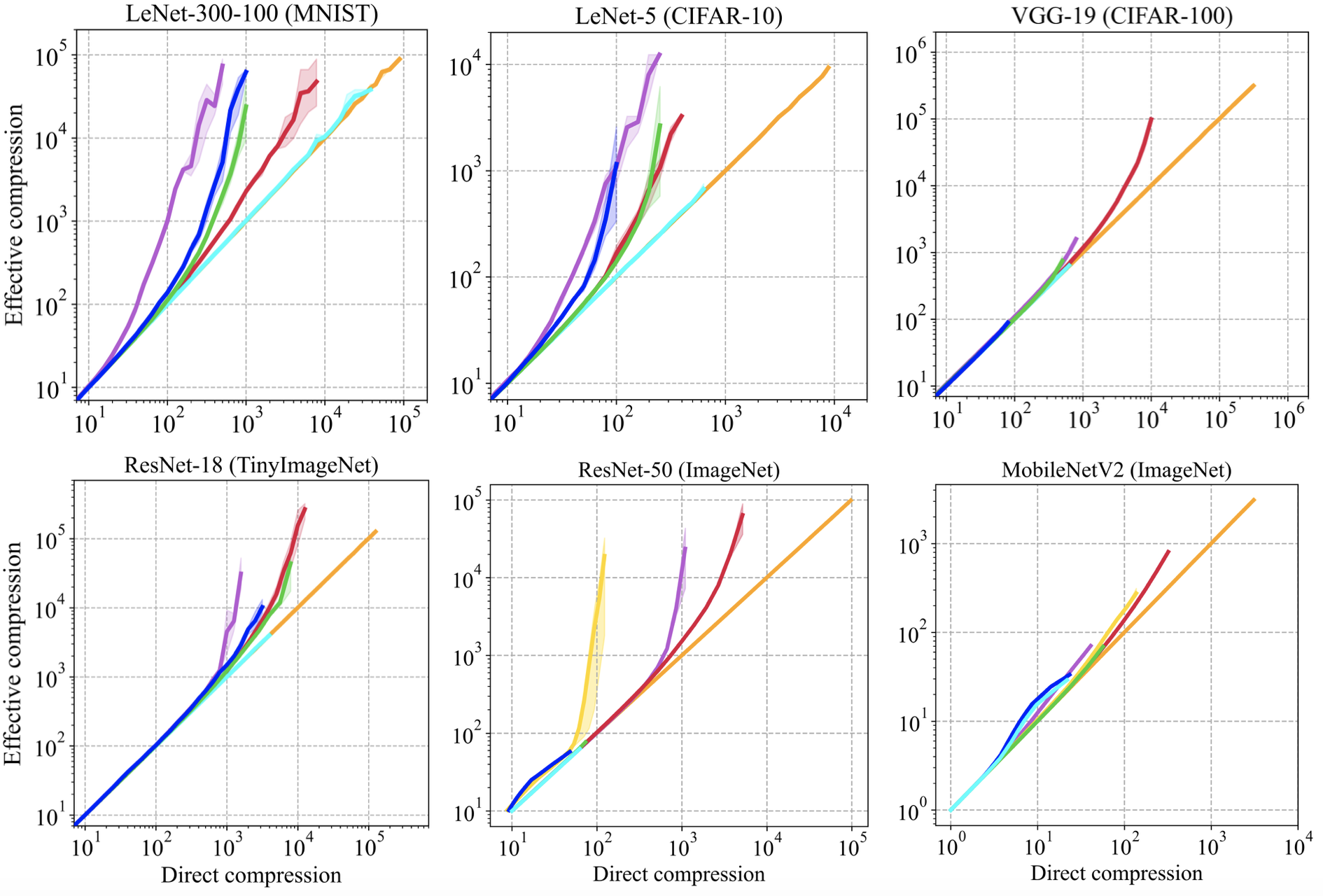}
\includegraphics[width=0.7\linewidth]{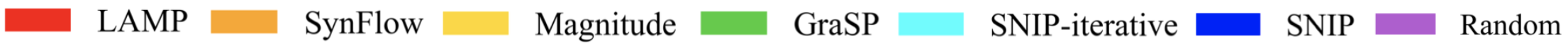}
\caption{Effective versus direct compression across different pruning methods and architectures (curves and bands represent min/average/max across 3 seeds where subnetworks disconnect last among a total of 5 seeds).}
\label{Fig:EffectiveSparsityA}
\end{figure}

\paragraph{Notation.} Consider an $L$-layer neural network $f(\mathbf{\Theta};x)$ with weight tensors $\mathbf{\Theta}=\{\Theta_{\ell}\}_{\ell=1}^{L}$ for $\ell\in[L]$. A subnetwork is specified by a set of binary masks that indicate unpruned parameters $M_{\ell}\in\{0,1\}^{|\Theta_{\ell}|}$. With $\mathbf{M}=\{M_{\ell}\}_{\ell=1}^{L}$, it is given by $f(\mathbf{\Theta}\odot\mathbf{M};x)$ where $\odot$ is Hadamard product. Note that biases and batchnorm parameters \citep{batchnorm} are normally considered unprunable. Direct sparsity, the fraction of pruned weights, is given by $s(\mathbf{M})=1-\sum_{\ell}\lVert M_{\ell}\rVert_0\big/\sum_{\ell}|M_{\ell}|$ and direct compression rate is defined as $(1-s(\mathbf{M}))^{-1}$.

\paragraph{Results.} Figure \ref{Fig:EffectiveSparsityA} reveals that different algorithms tend to develop varying amounts of inactive connections. For example, effective compression of subnetworks pruned by LAMP consistently reaches $10\times$ of their direct compression across all architectures, at which point at least nine in ten unpruned connections are effectively inactivated. Other methods (e.g., SNIP on VGG-19) remove entire layers early on, before any substantial differences between effective and direct compression emerge. Similarly, magnitude pruning applied after training disconnects most models very quickly and hence is not shown (e.g., LeNet-5 and VGG-19). In contrast, the picture is different with ResNet-50 whose residual connections might have allowed higher compression rates for this method. For example, we observe that the network is, in fact, two orders of magnitude less dense when direct sparsity reads just above $0.99$. SNIP-iterative and especially SynFlow demonstrate a truly unique property: subnetworks pruned by these two algorithms exhibit practically equal effective and direct compressions, and, in the case of SynFlow, disconnect only at very high compression rates. What makes them special? Both SynFlow and SNIP-iterative are multi-shot pruning algorithms that remove parameters over $100$ and $300$ iterations, respectively. SynFlow ranks connections by their $\ell_1$ path norm (sum of weighted paths passing through the edge, where the weight of a path is the product of magnitudes of weights of its edges). SNIP uses connection sensitivity scores from \citet{snip}  $|\frac{\partial\mathcal{L}}{\partial \theta_i}\theta_i|$ as a saliency measure where $\mathcal{L}$ is the loss function. Both these pruning criteria assign the lowest possible score of zero to inactive connections, scheduling them for immediate removal in the subsequent pruning iteration. Thus, by virtue of their iterative design, these two methods produce subnetworks with little to no difference between effective and direct compression. They are fortuitously designed to prune inactivated edges, which might explain their performance in high compressions.

\begin{figure}[ht]
\centering
\includegraphics[width=0.98\linewidth]{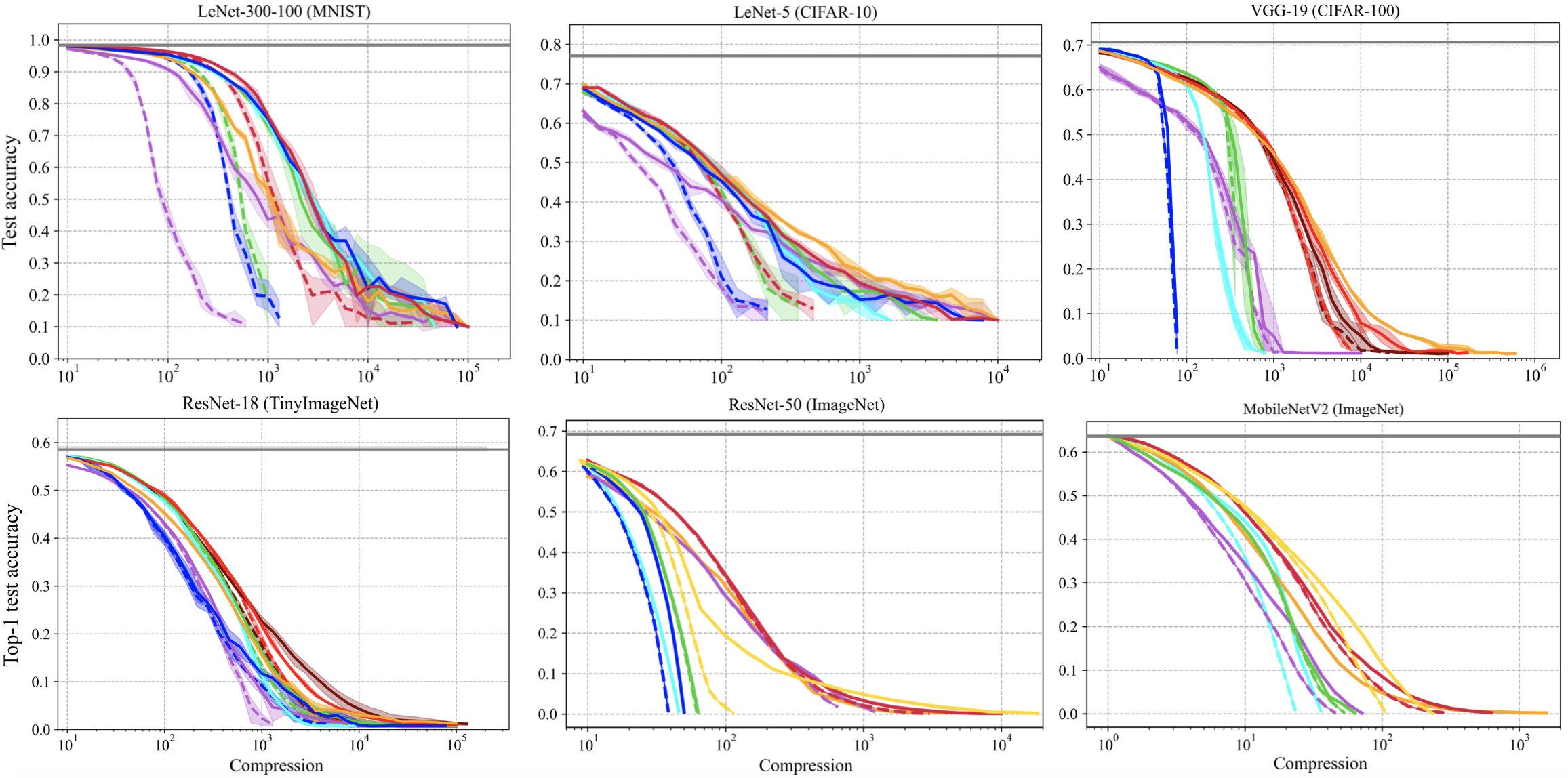}
\includegraphics[width=0.7\linewidth]{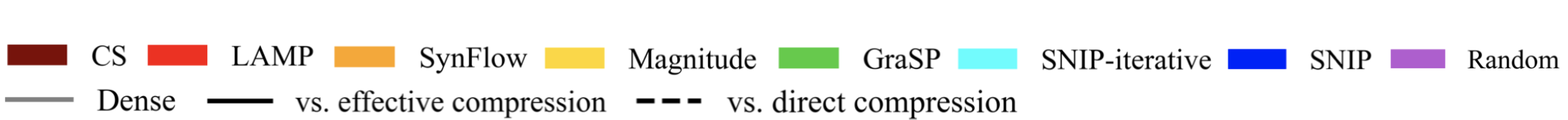}
\caption{Test accuracy (min/average/max) of subnetworks trained from scratch after being pruned by different algorithms plotted against direct (dashed) and effective (solid) compression. Dashed and solid curves overlap for SynFlow and SNIP-iterative. Solid curves are fitted to scatter data (not shown for clarity of the presentation) as in Figure \ref{Fig:2}.}
\label{Fig:EffectiveSparsityB}
\end{figure}

\citet{synflow} compare SynFlow to SNIP and GraSP using direct sparsity, claiming it vastly superior in high compression regimes. However, pruning methods that generate large amounts of inactivated connections are clearly at a significant disadvantage in the original direct framework. Figure \ref{Fig:EffectiveSparsityB} shows that the performance gap between SynFlow and other methods shrinks on all tested architectures under effective compression. The most dramatic changes are perhaps evident with LeNet-300-100 where SynFlow significantly dominates both SNIP and GraSP in direct comparison, but becomes strictly inferior when taken to the more telling effective compression. On the other hand, differences are not as pronounced on purely convolutional architectures such as VGG-19, and ResNet-18. Feature maps in convolutional layers are connected via groups of several parameters (kernels), making them more robust to inactivation compared to neurons in fully-connected layers.

\paragraph{Computing effective sparsity.} In advocating the use of effective sparsity, we must make sure that it can be calculated efficiently. We propose a simple approach leveraging SynFlow; note that a connection is inactive if and only if it is not part of any path from input to output. Assuming that unpruned weights are non-zero, this is equivalent to having zero $\ell_1$ path norm. \citet{synflow} observe that path norms can be efficiently computed with one pass on the all-ones input as $|\frac{\partial\mathcal{R}}{\partial\theta_i}\theta_i|$, where $\mathcal{R}=\mathbb{1}^{\top}f^{*}(|\mathbf{\Theta}|\odot\mathbf{M},\mathbb{1})$ and $f^{*}$ is the linearized version of the original network $f$. For deep models, rescaling of weights may be required to avoid numerical instability \citep{synflow}.

\section{Layerwise sparsity quotas (LSQ) and a novel allocation method (IGQ)}
\label{Sec:Pistons}
Inspired by \citet{missingthemark} and \citet{sanitychecks}, we wish to confirm that SNIP, GraSP, and SynFlow work no better than random pruning with corresponding layerwise sparsity allocation. While \citet{missingthemark} and \citet{sanitychecks} only considered moderate compression rates up to $100\times$ and used direct sparsity as a reference frame, we reconfirm their conjecture in the effective framework and test it across the entire compression spectrum. We generate and train two sets of subnetworks: $(i)$ pruned by either SNIP, GraSP, and SynFlow (\emph{original}), and $(ii)$ randomly pruned while preserving layerwise sparsity quotas provided by each of these three methods (\emph{random}).

\begin{figure}[h]
\centering
\includegraphics[width=0.85\linewidth]{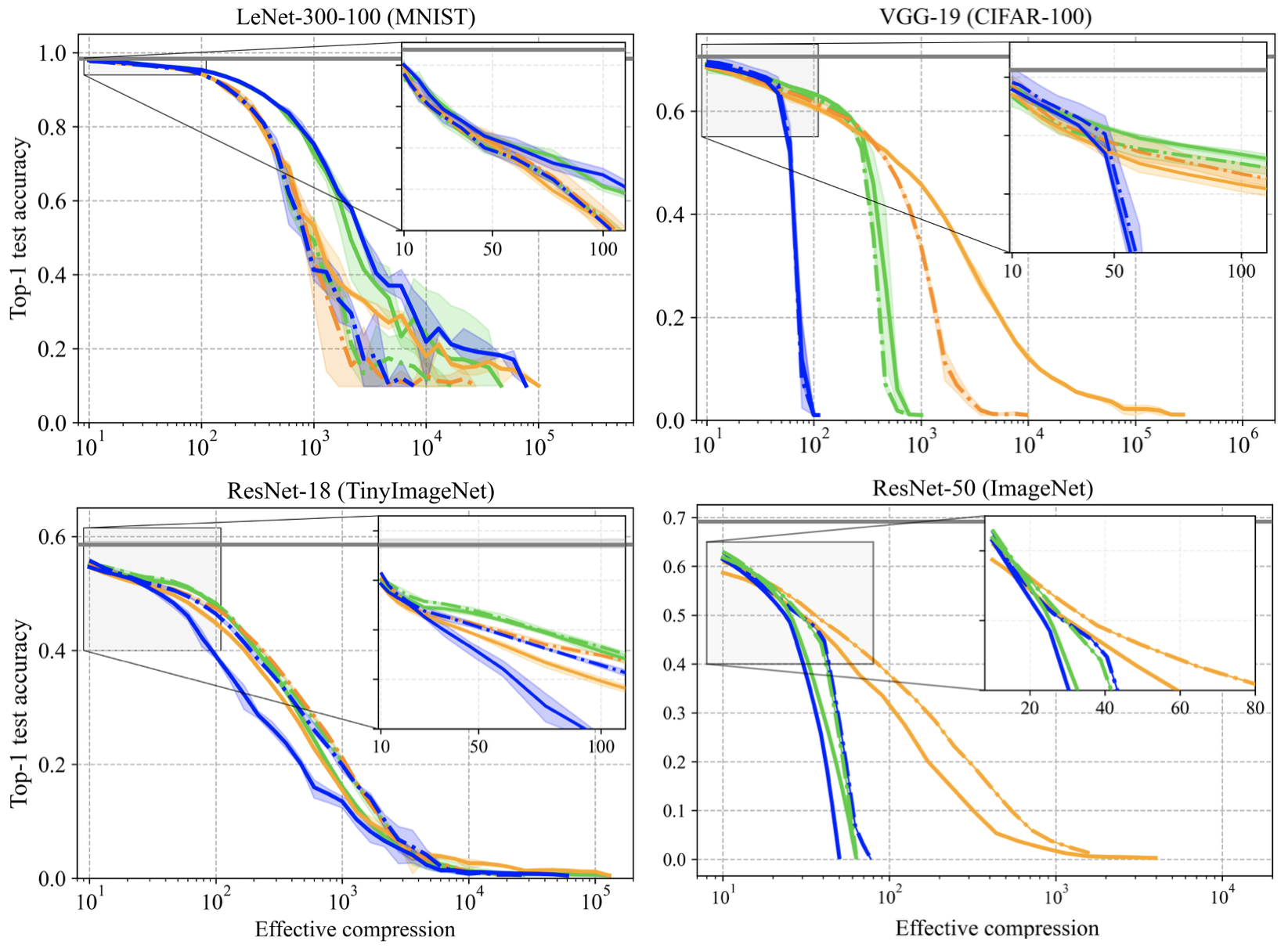}
\includegraphics[width=0.65\linewidth]{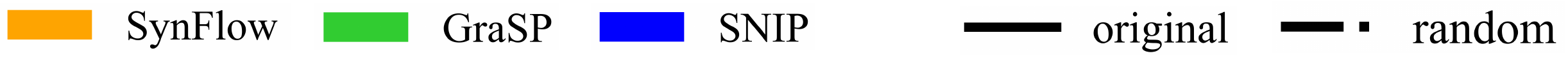}
\caption{Original methods for pruning at initialization (solid) and random pruning with corresponding layerwise sparsity quotas (dashdot). Test accuracy of the unpruned network is shown in grey.}
\label{Fig:MissingTheMark}
\end{figure}

\begin{figure}[H]
\centering
\includegraphics[width=0.9\linewidth]{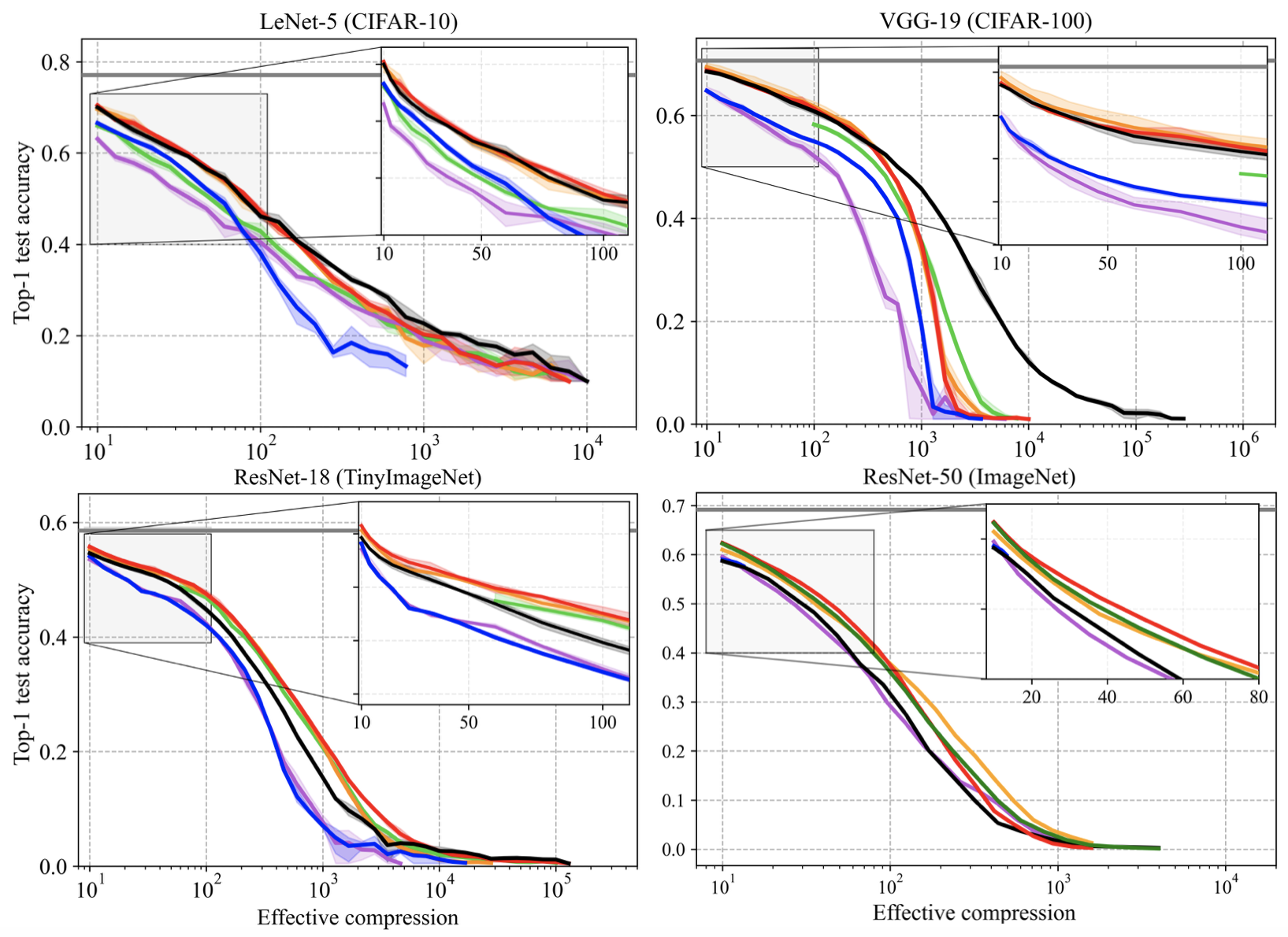}
\includegraphics[width=0.98\linewidth]{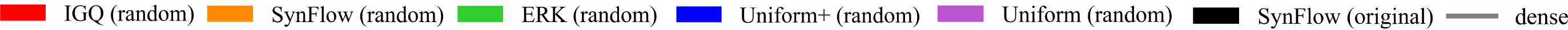}
\caption{Test performance of trained subnetworks after random pruning with different layerwise sparsity distributions. Original SynFlow (black) is shown for reference.}
\label{Fig:Random}
\end{figure}

Our results in Figure \ref{Fig:MissingTheMark} agree with observations made by \citet{missingthemark} and \citet{sanitychecks}: in the $10\times$--$100\times$ compression range, all three random pruning algorithms perform similarly (LeNet-300-100, VGG-19) or better (ResNet-18, ResNet-50) than their original counterparts. Effective sparsity allows us to faithfully examine higher compression, where the evidence is more equivocal. Similar patterns are still seen on ResNet-18; however, the original SNIP and GraSP beat random pruning with corresponding layerwise sparsities by a wide margin starting at $100\times$ compression on LeNet-300-100. Random pruning associated with SynFlow matches original SynFlow on the same network for longer, up to $1,000\times$ compression. On VGG-19, SynFlow bests the corresponding random pruning from about $500\times$  compression onward, while the original SNIP suffers from disconnection early on together with its random variant. Despite these nuances in the high compression regime, random pruning with specific layerwise sparsity quotas fares extremely well in the moderate sparsity regime (up to $99\%$) and is even competitive to full-fledged SynFlow (see Figure \ref{Fig:Random}). Therefore, random pruning can be a cheap and competitive alternative to more sophisticated and resource-consuming algorithms. This phenomenon is also reconfirmed in a recent study, which states that randomly pruned networks with carefully crafted LSQ can match the performance of their dense counterparts while comparing favorably in terms of adversarial robustness, out-of-distribution detection, and uncertainty estimation \citep{unreasonable}. In particular, they consider LSQ derived from SNIP and find it among the best performing sparsity distributions for random pruning. Alas, SNIP and other methods from Figure \ref{Fig:MissingTheMark} require expensive computations just to retrieve the corresponding pruning ratios, which may still suffer from issues like layer-collapse. This motivates us to ask: can we engineer readily computable and consistently well-performing sparsity quotas?

To our knowledge, there are only a few \emph{ab-initio} approaches in the literature to allocate sparsity in a principled fashion. \textit{Uniform} is the simplest solution that keeps sparsity constant across all layers. \citet{gale} 
give a modification (denoted \textit{Uniform+} following \citet{lamp}) that retains all parameters in the first convolutional layer and caps sparsity of the last fully-connected layer at $80\%$. A more sophisticated approach, \textit{Erd\"os-Renyi-Kernel (ERK)}, sets the density of a convolutional layer with kernel size $w\times h$, fan-in $n_{\text{in}}$ and fan-out $n_{\text{out}}$  proportional to $(w+h+n_{\text{in}}+n_{\text{out}})/(w\cdot h\cdot n_{\text{in}}\cdot n_{\text{out}})$. Although originally used as a sparsity distribution schema for methods with dynamic sparse structres (SET by \citet{set} and RigL by \citet{rigl}), we follow \citet{lamp} and use ERK as a baseline sparsity distribution for sparse-to-sparse training with a fixed subnetwork topology. The last two approaches are unable to support the entire range of sparsities: Uniform+ can only achieve moderate \emph{direct} compression because of the prunability constraints on its first and last layer, while both direct and effective sparsity levels achievable with ERK are often lower bounded. For example, the density of certain layers of VGG-16 set by ERK exceeds $1$ when cutting less than $99\%$ of parameters, unless excessive density is redistributed. \citet{sanitychecks} propose Smart-Ratios, which is an ad-hoc distribution method that requires the density of the $i$-th layer within an $L$-layer network to be proportional to $(L-l+1)^2+(L-l+1)$. This method was developed exclusively for VGG-like networks and, like ERK and Uniform$+$, can be infeasible for certain sparsities.

To avoid problems that riddle Uniform+, ERK, and smart-ratios, we require that any layerwise sparsity quotas must be attainable for any level of network sparsity $s\in[0,1]$. At the same time, neither layer should be removed in its entirety unless $s=1$ to avoid layer-collapse inherent to SNIP and some other global pruning methods. These requirements lead us to formulate a formal definition for layerwise sparsity quotas to guide principled future research into sparsity allocation.

\paragraph{Definition 1 (Layerwise Sparsity Quotas).}
\textit{ A function $\mathcal{Q}\colon [0,1]\rightarrow [0,1]^{L}$ mapping a target sparsity  $s$ to layerwise sparsities  $\{s_{\ell}\}_{\ell=1}^{L}$ is called Layerwise Sparsity Quotas (LSQ) if it satisfies the following properties: (i) total sparsity: for any $s\in[0,1]$, $s\sum_{\ell}|\Theta_{\ell}|=\sum_{\ell}s_{\ell}|\Theta_{\ell}|$,
and (ii) layer integrity: for all layers $\ell\in[L]$, $[\mathcal{Q}(s)]_{\ell}<1$ if $s<1$.}

\begin{wrapfigure}{r}{0.5\textwidth}
\centering
\includegraphics[width=0.78\linewidth]{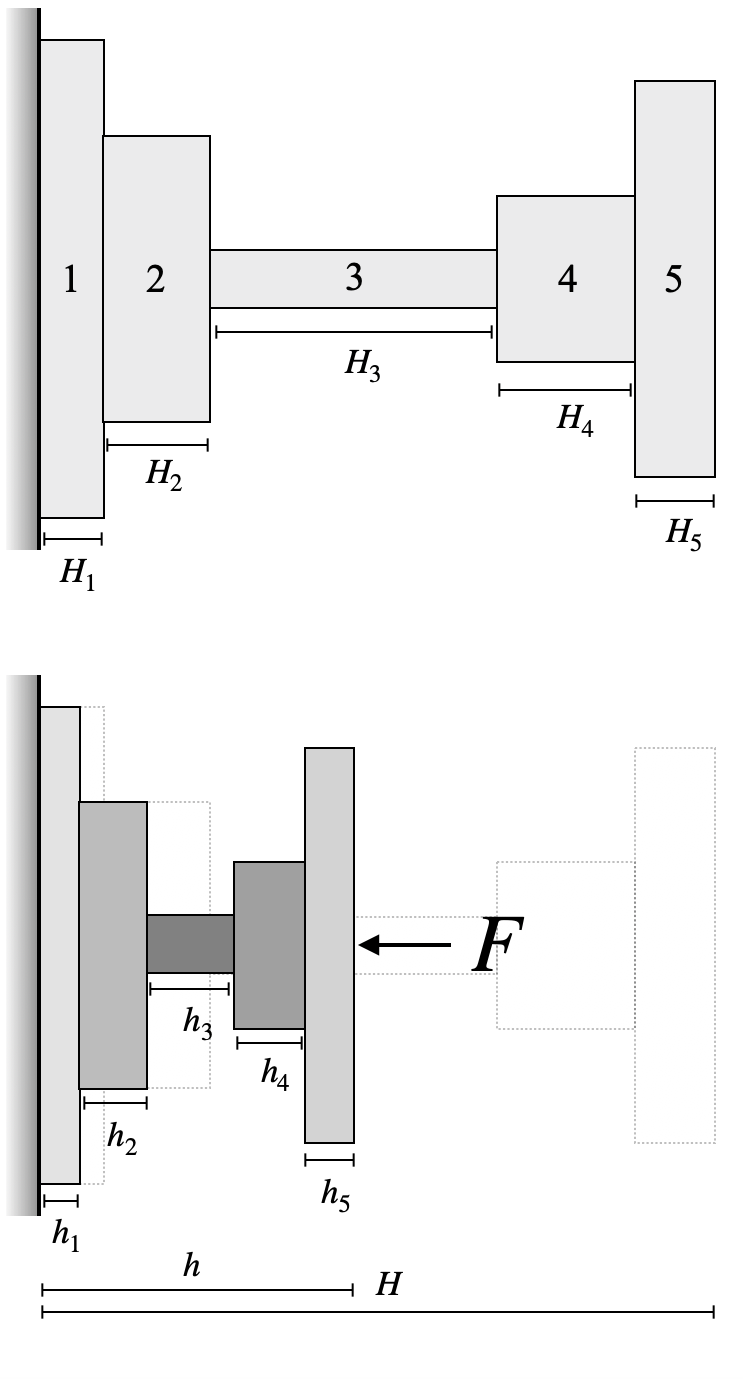}
\caption{Schematic diagram of the cylinder system underlying IGQ compression of LeNet-5. Each of the five layers of the network is represented by a cylinder of unit volume and height
 $H_{\ell}$ proportional to the number of parameters $|\Theta_{\ell}|$ in that layer. As force $F$ is applied to the outermost cylinder (5), the system transforms according to the Ideal Gas Law, yielding IGQ compression rates of $H_{\ell}/h_{\ell}$, while the overall network compression is $H/h = \sum_{{\ell}=1}^5H_{\ell}/\sum_{{\ell}=1}^5h_{\ell}$. Darker colors indicate higher compression. Note that cylinders are not drawn to scale.}
 \vspace{-10px}
\label{Fig:IGQSchema}
\end{wrapfigure}

\begin{figure}[H]
\centering
\includegraphics[width=0.99\linewidth]{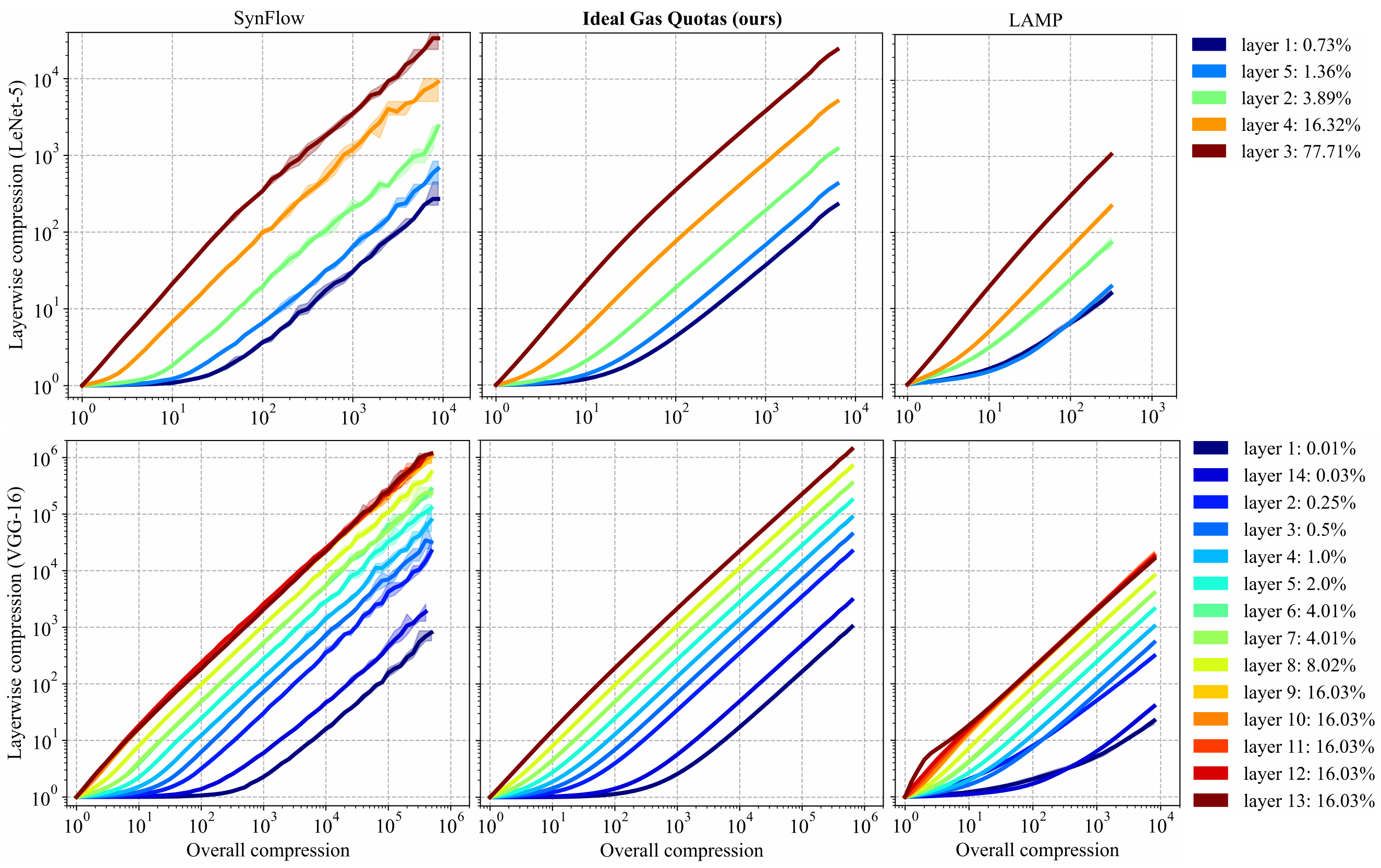}
\caption{Layerwise direct compression quotas of LeNet-5 (top) and VGG-16 (bottom) associated with SynFlow (left), our IGQ (middle), and LAMP (right). Percentages indicate layer sizes relative to the total number of parameters; colors are assigned accordingly from blue (smaller layers) to red (larger layers). Curves of LAMP and SynFlow end when the underlying network disconnects.}
\label{Fig:LayerwiseCompression}
\end{figure}

\begin{figure}[H]
\centering
\includegraphics[width=0.9\linewidth]{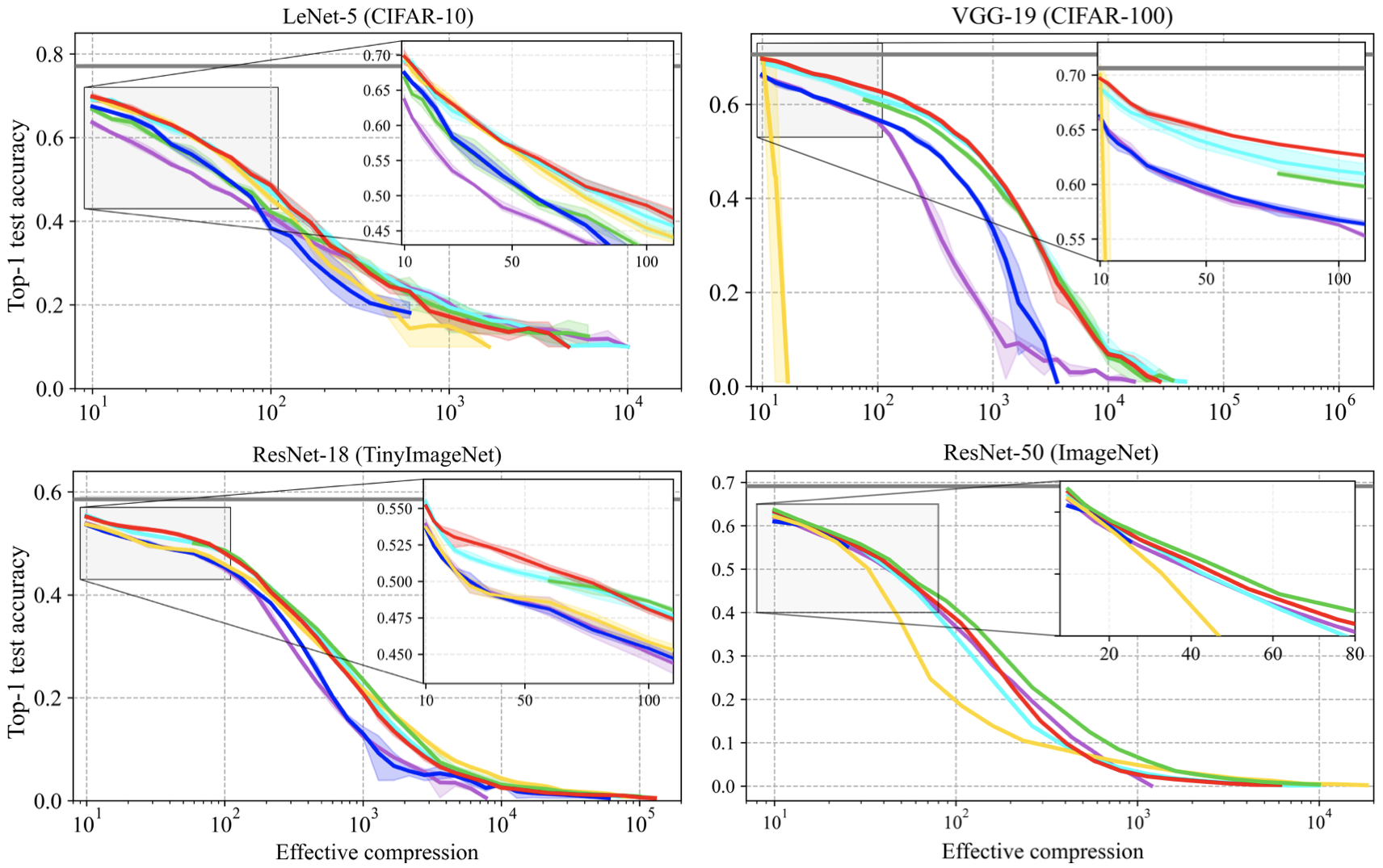}
\includegraphics[width=0.75\linewidth]{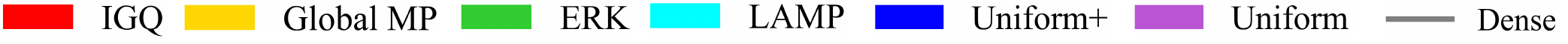}
\caption{Test performance of retrained subnetworks after magnitude-based pruning. Uniform+ is not shown for LeNet-300-100 since it is designed for convolutional networks.}
\label{Fig:Magnitude}
\end{figure}

\paragraph{Ideal Gas Quotas (IGQ).}Aiming to unfold the secret of well-performing layerwise compression quotas associated with such global pruning algorithms as SNIP, LAMP, and SynFlow, we note that they prune larger, parameter-heavy layers more aggressively than smaller layers (Figure \ref{Fig:LayerwiseCompression}), which has been already conjectured to be a desirable property \citep{sanitychecks}. To design a valid LSQ with this feature, we consult an intuitive (although lacking formal connection with neural network pruning) analogy from physics. In particular, we interpret compression of a multi-layer network as compression of stacked gas-filled weightless cylinders of unit volume and height equal to the size of the corresponding layer (Figure \ref{Fig:IGQSchema}). As force is applied to the system, the Ideal Gas Law governs the compression rate of each cylinder, giving the final compression distribution which we interpret as the layerwise compression (sparsity) distribution within the given network. Using simple algebra, we arrive at compression quotas $\{F|\Theta_{\ell}|+1\}_{\ell=1}^{L}$ (or sparsity quotas $\{1-(F|\Theta_{\ell}|+1)^{-1}\}_{\ell=1}^{L}$) parameterized by the force $F$ that controls the overall sparsity of the network. Thus, we selected cylinder dimensions to encode our prior belief that larger layers can withstand higher pruning rates since ``flatter'' cylinders undergo lighter compression under the same external force (compression constraint). Given a target sparsity $s$, the needed value of $F$ can be instantly found using binary search to any given precision.  IGQ clearly satisfies all requirements of Definition 1 and applies higher compression to larger layers, as desired. In principle, IGQ is applicable in a variety of contexts with use-cases in pruning before training (in conjunction with random pruning), during training (e.g., as default LSQ for RigL \citep{rigl}), and after training (e.g., together with magnitude pruning). In this study, we adopt the first and the last scenarios to evaluate IGQ against baselines (Figures \ref{Fig:Random}, \ref{Fig:Magnitude}).

\paragraph{Random pruning with IGQ.} While \citet{unreasonable} experiment with lower sparsities (up to $90\%$) and a slightly different set of LSQ, our results largely match their evidence. In particular, we also find that ERK consistently outperforms more naive baselines like Uniform and Uniform+. Although ERK sometimes exhibits similar (ResNet-18) or even better (VGG-19 compressed to $1,000\times$ or higher) performance than IGQ, it yields invalid layerwise sparsity quotas when removing less than $98\%$ and $99\%$ of parameters from ResNet-18 and VGG-19, respectively, thus failing to satisfy Definition 1. Uniform$+$ produces invalid layerwise compressions from $40\times$ onward for ResNet-50. In the moderate sparsity regime (up to $99\%$), subnetworks pruned by IGQ reach unparalleled performance after training, especially on ResNet-50. Across all architectures, random pruning with IGQ and SynFlow sparsity quotas are almost indistinguishable from each other, suggesting that IGQ successfully mimics the quotas produced by SynFlow, which require substantial effort to compute. Therefore, judging by a tripartite criterion of test performance, compliance with Definition 1, and computational efficiency, IGQ beats all baselines.

\paragraph{Magnitude pruning with IGQ.}In the second set of experiments, we pretrain fully-dense models and prune them by magnitude using global methods (Global Magnitude Pruning, LAMP) or layer-by-layer respecting sparsity allocation quotas (Uniform, Uniform+, ERK, and IGQ). Then, we revert the unpruned weights back to their original random values and fully retrain the resulting subnetworks to convergence. Results are displayed in Figure \ref{Fig:Magnitude} in the framework of effective compression. Overall, our method for distributing sparsity in the context of magnitude pruning performs consistently well across all architectures and favorably compares to other baselines, especially in moderate compression regimes of $100\times$ or less. Even though Global magnitude pruning can marginally outperform IGQ, it is completely unreliable on VGG-19. ERK appears slightly better than IGQ on VGG-19, ResNet-18 and ResNet-50 at extreme sparsities, however, it performs much worse on LeNet-5 and has other general deficiencies as discussed earlier. Another close rival of IGQ is LAMP, which performs very similarly but is still unable to reach its performance on VGG-19, ResNet-18 and ResNet-50 in moderate compression regimes. Note, however, that all presented methods require practically equal compute and time; thus, the evidence in Figure \ref{Fig:Magnitude} is not meant to advertise IGQ as a cheaper alternative to LAMP but rather to illustrate the effectiveness of IGQ.

\section{Effective pruning}
\label{Sec:Pruner}
\begin{wrapfigure}{r}{0.5\textwidth}
\vspace{-10pt}
\begin{center}
\includegraphics[width=0.5\textwidth]{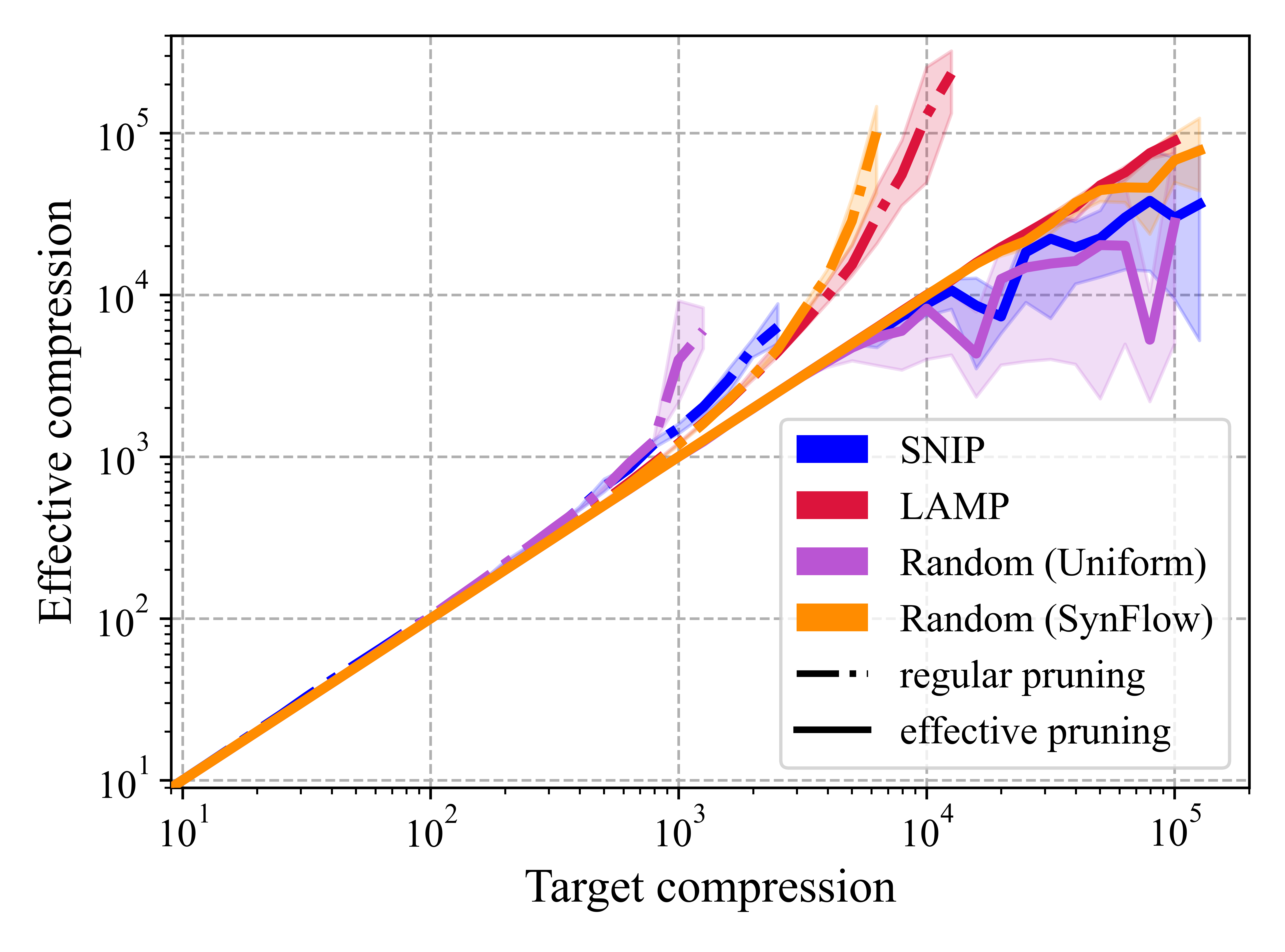}
\end{center}
\caption{Effective compression produced by regular (dashdot) and our effective (solid) pruning on ResNet-18 according to ranking-based (left) and random (right) algorithms. Our procedures help pruning reach target effective sparsity, falling short only when the subnetwork is on the brink of disconnection.}
\label{Fig:AEP1}
\end{wrapfigure}

Unlike pruning to a target direct sparsity, pruning to achieve a particular \emph{effective} sparsity can be tricky. Here, we present an extension to algorithms for pruning at initialization or after training that achieves this goal efficiently, when possible (see Figure \ref{Fig:AEP1}).

\paragraph{Effective ranking-based pruning.} Algorithms like GraSP, SynFlow, and LAMP rank parameters by some notion of importance to guide pruning.  When such a ranking $R\colon\mathbf{\Theta}\rightarrow\mathbb{R}$ is available, we employ binary search for the appropriate cut-off threshold $t$ in $\mathcal{O}(\log |\mathbf{\Theta}|)$ time. This approach leverages the following monotonicity property: given two pruning thresholds $t_1,t_2\in R$ and corresponding subnetworks $S_1,S_2$, we have $t_1\leq t_2$ if and only if $S_2\subseteq S_1$, which implies $\text{EffectiveSparsity}(S_1)\leq\text{EffectiveSparsity}(S_2)$ (note that in general $\text{Sparsity}(S_1)\leq\text{Sparsity}(S_2)$ does not imply the last inequality above). Thus, binary search will branch in the correct direction.


\paragraph{Effective random pruning.} In Section \ref{Sec:Pistons}, we saw that random pruning with carefully crafted layerwise sparsity quotas $\mathcal{Q}\colon[0,1]\rightarrow[0,1]^{L}$ fares well (especially in the framework of effective sparsity) with more sophisticated pruning methods, proving to be a cheaper and simpler alternative. Effective pruning without parameter scores is more challenging because there is no obvious way to produce a neat chain of embedded subnetworks as above. For example, given two subnetworks $S_1$ and $S_2$, $\text{Sparsity}(S_1)\leq\text{Sparsity}(S_2)$ does not imply $\text{EffectiveSparsity}(S_1)\leq\text{EffectiveSparsity}(S_2)$. Assigning random scores requires $\mathcal{O}(|\mathbf{\Theta}|)$ time to ensure that any cut-off threshold yields LSQ according to $\mathcal{Q}$, which is not scalable.


To circumvent this issue, we design an improved algorithm that produces embedded subnetworks on each iteration, allowing binary search to work (see Algorithm \ref{Alg:AERP}). Starting from the extreme subnetworks $S_1$ (fully-dense, corresponding to masks $\mathbf{M}^{(1)}$) and $S_2$ (fully-sparse, corresponding to masks $\mathbf{M}^{(2)}$), we narrow the sparsity gap between them while preserving $S_2\subseteq S_1$ so that $\text{EffectiveSparsity}(S_1)\leq\text{EffectiveSparsity}(S_2)$. For each layer, we keep track of unpruned connections $U_{\ell}$ of $S_1$ and pruned connections $P_{\ell}$ of $S_2$, randomly sample parameters $T_{\ell}$ from $U_{\ell}\cap P_{\ell}$ according to $\mathcal{Q}$ and form another network $S$ by pruning out $\bigcup_{\ell}T_{\ell}$ from $S_1$ (or, equivalently, reviving in $S_2$). Depending on where effective sparsity of $S$ lands relative to target $s$, we assign $S$ to either $S_1$ or $S_2$ and branch. Since connections to be pruned from $S_1$ (or revived in $S_2$) are chosen randomly at each step, weights within the same layer have equal probability of being pruned. Once $S_1$ and $S_2$ are only $1$ parameter away from each other, the algorithm returns $S_1$, yielding a connected model. Note that this algorithm implicitly requires the LSQ function $\mathcal{Q}$ to be layerwise monotone: if $s_1\leq s_2$, then $[\mathcal{Q}(s_1)]_{\ell}\leq[\mathcal{Q}(s_2)]_{\ell}$ for each layer $\ell\in[L]$. This is a reasonable assumption and is satisfied in practice (see Figure \ref{Fig:LayerwiseCompression}).

\section{Discussion}
\label{Sec:Discussion}
In our work, we argue that \emph{effective sparsity (effective compression)} is the correct benchmarking measure for pruning algorithms since it discards effectively inactive connections and represents the true remaining connectivity pattern. Moreover, effective sparsity allows us to study extreme compression regimes for subnetworks that otherwise appear disconnected at much lower direct sparsities. We initiate the study of current pruning algorithms in this refined frame of reference and rectify previous benchmarks. To facilitate the use of effective sparsity in future research, we describe low-cost procedures to both compute and achieve desired effective sparsity when pruning. Lastly, with effective sparsity allowing us to zoom more fairly into higher compression regimes than previously possible, we examine random pruning with prescribed layerwise sparsities and propose our own readily computable quotas (IGQ) after establishing conditions reasonable LSQ should fulfill. We show that IGQ, while allowing for any level of sparsity, is more advantageous than all existing similar baselines (Uniform, ERK) and gives comparable performance to sparsity quotas derived from more sophisticated and computationally expensive algorithms like SynFlow.

\label{AppendixAlgorithm}
\begin{algorithm}[t]
\SetAlgoLined
\SetKwInOut{Input}{Input}
\Input{ Desired effective sparsity $s$; LSQ function $\mathcal{Q}\colon[0,1]\rightarrow[0,1]^{L}$.}
 $i\leftarrow 0$; $j\leftarrow |\mathbf{\Theta}|$;
 $\mathbf{M}^{(1)}\leftarrow\mathbf{1}$; $\mathbf{M}^{(2)}\leftarrow\mathbf{0}$; $P_{\ell},U_{\ell}\leftarrow \Theta_{\ell}$ for all $\ell\in[L]$\;
 \While{$j-i>1$}{
   $m\leftarrow \floor{(i+j)/2}$; $\{s_{\ell}\}_{\ell=1}^{L}\leftarrow\mathcal{Q}(m/|\mathbf{\Theta}|)$\;
 \For{$\ell\in[L]$}{
 $\text{CurrSparsity}_{\ell}\leftarrow (1-|U_{\ell}|/|\Theta_{\ell}|)$\;
  $T_{\ell}\leftarrow\text{RandomSelect}(\text{from}=U_{\ell}\cap P_{\ell},\text{size}=|\Theta_{\ell}|(s_{\ell}-\text{CurrSparsity}_{\ell}))$\;
  $M_{\ell}\leftarrow\text{CreateMask}(\text{pruned}=\Theta_{\ell}\setminus[U_{\ell}\setminus T_{\ell}],\text{unpruned}=U_{\ell}\setminus T_{\ell})$\;}
  $\mathbf{M}\leftarrow\{M_{\ell}\}_{\ell=1}^{L}$\;
  \eIf{$\text{EffectiveSparsity}(\mathbf{M})<s$}{
   $U_{\ell}\leftarrow U_{\ell}\setminus T_{\ell}$ for all $\ell\in[L]$; $\mathbf{M}^{(1)}\leftarrow\mathbf{M}$; $i\leftarrow m$\;
   }{
   $P_{\ell}\leftarrow P_{\ell}\setminus T_{\ell}$ for all $\ell\in[L]$; $\mathbf{M}^{(2)}\leftarrow\mathbf{M}$; $j\leftarrow m$\;
 }
 }
 \SetKwInOut{Return}{Return}
 \Return{ Masks $\mathbf{M}^{(1)}$ s.t.
$\text{EffectiveSparsity}(\mathbf{M}^{(1)})\sim s$, $\lVert M^{(1)}_{\ell}\rVert_0=|\Theta_{\ell}|(1-[\mathcal{Q}(s)]_{\ell})$.}
\caption{Approximate Effective Random Pruning}
\label{Alg:AERP}
\end{algorithm}

\paragraph{Limitations and Broader Impacts.} We hope that the lens of effective compression will spur more research in high compression regimes. One possible limitation is that it is harder to control effective compression exactly. In particular, using different seeds might lead to slightly different effective compression rates. However, these perturbations are minor. Additionally, one might argue that for some architectures accuracy drops precipitously with higher compression thus making very sparse subnetworks less practical. We hope that opening the study of high compressions will allow to explore how to use sparse networks as building blocks, for instance using the power of ensembling \citep{ensembling}. Our framework allows a principled study of this regime. Finally, since effective compression strips away unnecessary computational units, it offers a potentially higher resource efficiency during both inference and training without compromising the flexibility of unstructured pruning or requiring specialized hardware \citep{hardware}.

\acks{Both authors were supported by the National Science Foundation under NSF Award 1922658. Neither of the authors has any competing interests to report.}

\appendix
\section{Experimental details}
\label{Sec:AppendixHyperparameters}
Our experimental work encompasses seven different architecture-dataset combinations: LeNet-300-100 \citep{lenets} on MNIST (Creative Commons Attribution-Share Alike 3.0 license), LeNet-5 \citep{lenets} and VGG-16 \citep{vggnets} on CIFAR-10 (MIT license), VGG-19 \citep{vggnets} on CIFAR-100 (MIT license), and ResNet-18 \citep{resnet} on TinyImageNet (MIT license), ResNet-50 and MobileNetV2 \citep{mobilenets} on ImageNet-2012 \citep{imagenet}. Following \citet{missingthemark}, we do not reinitialize subnetworks after pruning (we revert back to the original initialization when pruning a pretrained model by LAMP). We use our own implementation of all pruning algorithms in TensorFlow except for GraSP, for which we use the original code in PyTorch published by \citet{grasp}. All non-ImageNet runs were repeated 3 times for stability of results. Training was performed on an internal cluster equipped with NVIDIA RTX-8000, NVIDIA V-100, and AMD MI50 GPUs. Hyperparameters and training schedules used in our experiments are adopted from related works and are listed in Table \ref{Table:1}. We apply standard augmentations to images during training. In particular, we normalize examples per-channel for all datasets and randomly apply: $(i)$ shifts by at most 4 pixels in any direction and horizontal flips (CIFAR-10, CIFAR-100, and TinyImageNet), $(ii)$ rotations by up to 4 degrees (MNIST), and $(iii)$ random $224\times224$ crops and horizontal flips (ImageNet).

\begin{table}[H]
\centering
\begin{tabular}{llllllll}
\toprule
Model   & Epochs   & Drop epochs   & Batch  & LR   & Decay   & Source \\
\midrule
LeNet-300-100   & $160$ & $41/83/125$ & $100$ & $0.1$ & $5e$-$4$ & \citet{snip} \\
LeNet-5   & $307$   & $76/153/230$ & $128$ & $0.1$ & $5e$-$4$ & \citet{snip} \\
VGG-16   & $160$ & $80/120$ & $128$   & $0.1$ & $1e$-$4$ & \citet{missingthemark} \\
VGG-19   & $160$ & $80/120$ & $128$   & $0.1$ & $5e$-$4$ & \citet{grasp} \\
ResNet-18   & $200$ & $100/150$ & $256$ & $0.2$ & $1e$-$4$ & \citet{missingthemark} \\
ResNet-50   & $90$ & $30/60/80$ & $512$ & $0.4$ & $1e$-$4$ & \citet{missingthemark} \\
MobileNetV2   & $90$ & $30/60/80$ & $512$ & $0.4$ & $1e$-$4$ & \citet{missingthemark}\\
\bottomrule
\end{tabular}
\caption{Summary of experimental work. All architectures include batch normalization layers followed by ReLU activations. Models are initialized using Kaiming normal scheme (fan-avg) and optimized by SGD (momentum $0.9$) with a stepwise LR schedule ($10\times$ drop factor applied on specified \emph{drop epochs}). The categorical cross-entropy loss function is used for all models.}
\label{Table:1}
\end{table}

\section{Experiments with VGG-16}
\label{Sec:AppendixVGG16}
\begin{figure}[H]
\centering
\includegraphics[width=0.45\linewidth]{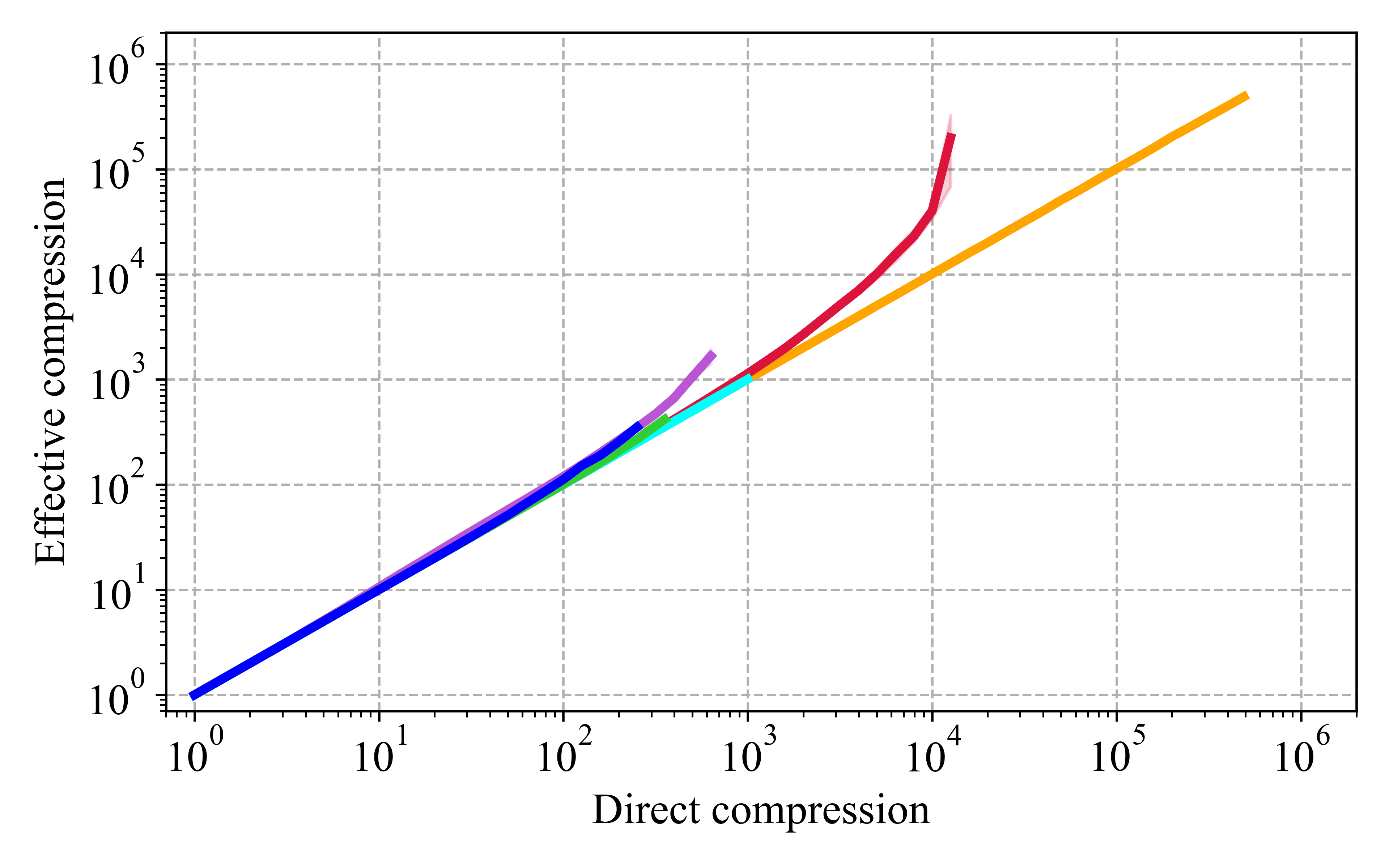}
\includegraphics[width=0.45\linewidth]{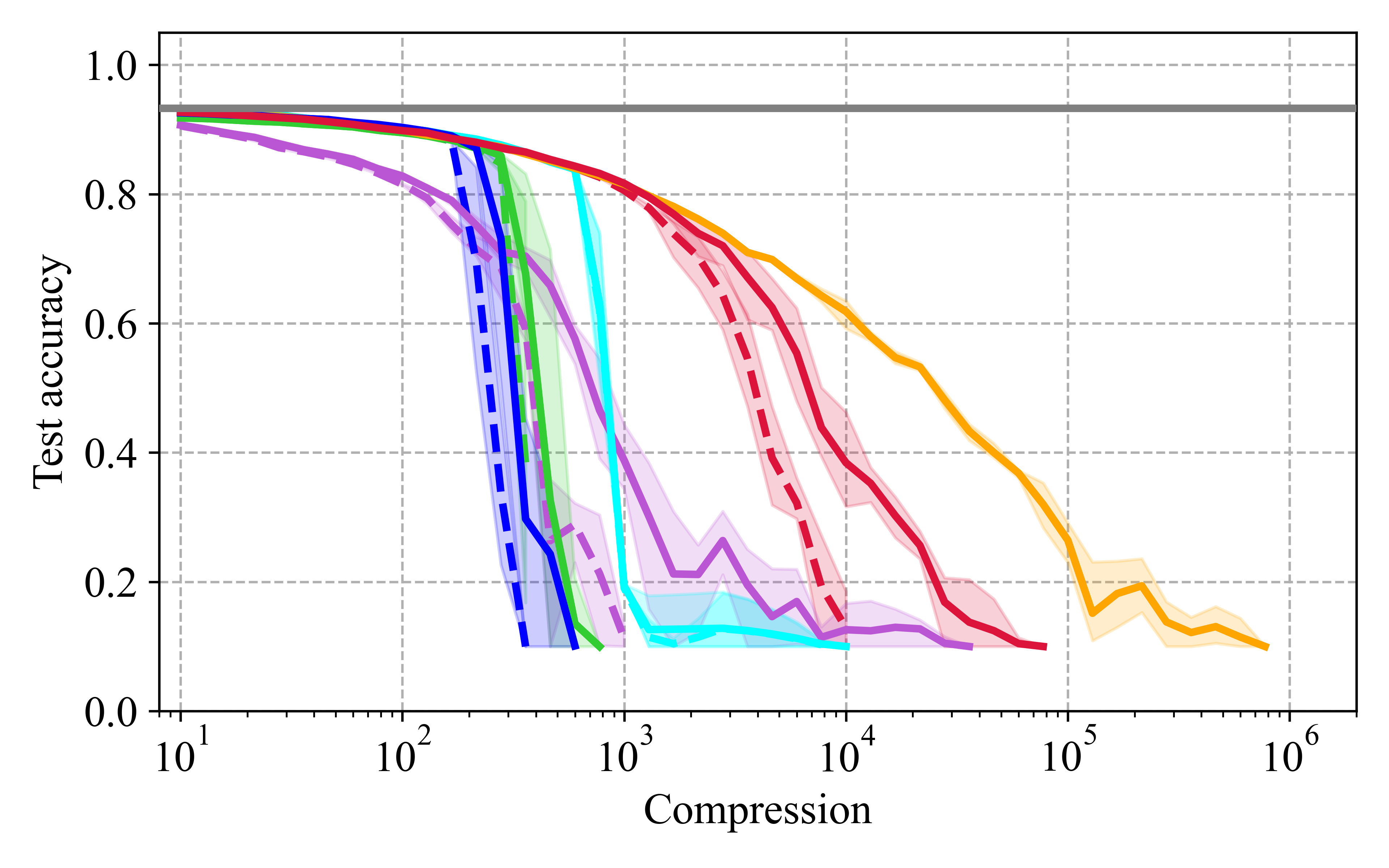}\\
\includegraphics[width=0.65\linewidth]{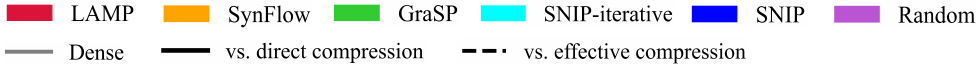}
\caption{Left: effective versus direct compression of VGG-16 when pruned by different algorithms. Right: test accuracy (min/average/max) of VGG-16 trained from scratch after being pruned by different algorithms plotted against direct (dashed) and effective (solid) compression. Dashed and solid curves overlap for SynFlow and SNIP-iterative.}
\label{Fig:EffectiveSparsityC}
\end{figure}

In Figure \ref{Fig:EffectiveSparsityC}, we display the results of our experiments with VGG-16 on CIFAR-10. As we argued in Section \ref{Sec:EffectiveSparsity}, higher sparsities are required for purely convolutional architectures (such as VGG-16) to develop inactive connections since feature maps are harder to disconnect. At the same time, several algorithms (SNIP, SNIP-iterative, GraSP) suffer from layer-collapse at modest sparsities ($99.9\%$ or less) and, hence, fail to develop significant amounts of inactive parameters. For this reason, as evident from Figures \ref{Fig:EffectiveSparsityA}, \ref{Fig:EffectiveSparsityB}, and \ref{Fig:EffectiveSparsityC}, VGG-16 arguably showcases the least differences between effective and direct compression among all tested architectures.

\vskip 0.2in
\bibliography{22-0415}

\end{document}